\documentclass[conference,a4paper,twocolumn]{IEEEtran}

\IEEEoverridecommandlockouts

\usepackage{booktabs}
\usepackage{multirow}

\ifCLASSINFOpdf
\else
\fi
  \usepackage{tipa}
  \usepackage[pdftex]{graphicx}
	\usepackage{epsfig}
	\usepackage{epstopdf}

\usepackage[cmex10]{amsmath}
\usepackage{subfigure}
\begin{document}



\title{Towards Improving the Performance of Pre-Trained Speech Models for Low-Resource Languages Through Lateral Inhibition}


\author{
\IEEEauthorblockN{
Andrei-Marius Avram$^{*}$, Răzvan-Alexandru Smădu$^{*}$, Vasile Păiș$^{{\dag}}$, \\ Dumitru-Clementin Cercel$^{*}$, Radu Ion$^{{\dag}}$, and Dan Tufiș$^{{\dag}}$
}
\IEEEauthorblockA{
$^{*}$Faculty of Automatic Control and Computers, University Politehnica of Bucharest, Romania\\
$^{\dag}$Research Institute for Artificial Intelligence  “Mihai Drăgănescu”, Romanian Academy\\
\{andrei\_marius.avram, razvan.smadu\}@stud.acs.upb.ro, dumitru.cercel@upb.ro, \{vasile,radu,tufis\}@racai.ro
}
}


%


\maketitle


\begin{abstract}
With the rise of bidirectional encoder representations from Transformer models in natural language processing, the speech community has adopted some of their development methodologies. Therefore, the Wav2Vec models were introduced to reduce the data required to obtain state-of-the-art results. This work leverages this knowledge and improves the performance of the pre-trained speech models by simply replacing the fine-tuning dense layer with a lateral inhibition layer inspired by the biological process. Our experiments on Romanian, a low-resource language, show an average improvement of 12.5\% word error rate (WER) using the lateral inhibition layer. In addition,  we obtain state-of-the-art results on both the Romanian Speech Corpus and the Robin Technical Acquisition Corpus with 1.78\% WER and 29.64\% WER, respectively.
\end{abstract}


\begin{IEEEkeywords}
Lateral Inhibition; Romanian Language; Speech Recognition; Wav2Vec 2.0
\end{IEEEkeywords}

%
\IEEEpeerreviewmaketitle

\section{Introduction}

Deep neural networks benefit from large amounts of annotated training data. However, annotated data is challenging to obtain in many settings. Except for English, generating thousands of hours of transcribed audio necessary to train a state-of-the-art speech recognition system is infeasible for most languages worldwide. Self-supervised learning~\cite{bao2021beit} has become the de facto technique for addressing this issue by first teaching a general data representation from unlabeled samples and then transferring the accumulated knowledge to a downstream task via fine-tuning~\cite{kenton2019bert}.

Working with self-supervision on unlabeled speech signals involves similar challenges as in computer vision. However, the research community continued to build pre-trained models on audio that have pushed further the state of the art in speech recognition. Schneider et al. \cite{schneider2019wav2vec}  introduced the Wav2Vec model, which encodes the input audio data into a latent space to create a contextualized representation employing a Transformer encoder \cite{vaswani2017attention}. Baevski et al.~\cite{baevski2020wav2vec} built Wav2Vec 2.0 on top of the previous work, mainly using the same model architecture while changing the pre-training objective to a discretized contrastive loss similar to the masked language model strategy from natural language processing.

Introduced by Păiș \cite{pais2022lateralinhib}, the lateral inhibition layer helps the model to learn when the annotated data is scarce. This paper investigates its application in transcribing human voice from audio files by integrating the lateral inhibition mechanism into a pre-trained automatic speech recognition (ASR) system. We choose the Wav2Vec 2.0 Base model pre-trained on 100k hours of unlabeled audio data extracted from VoxPopuli (i.e., Wav2Vec2.0-VP-100k)~\cite{wang2021voxpopuli}. We run our experiments on a low-resource language, namely the Romanian language. 

Our results for the experimental setup with the lateral inhibition layer show an average performance of 12.5\% word error rate (WER) on various dataset settings compared with the feed-forward layer.
In addition, we obtain state-of-the-art results on the Romanian Speech Corpus (RSC) \cite{georgescu2020rsc}  with 1.78\% WER, using fewer training data than the previous model, and on the Robin Technical Acquisition Speech Corpus (RTASC) \cite{puaics2021human} with 29.64\% WER, using the same training data.

We can summarize our main contributions as follows:
(i) applying the technique of neural lateral inhibition to ASR;
(ii) performing an analysis of the improvements brought by the lateral inhibition layer; 
(iii) to the best of our knowledge, creating the first publicly available Romanian Wav2Vec 2.0 model\footnote{https://huggingface.co/racai} (called  RoWav2Vec2.0-VP-100k-LI) that was thoroughly evaluated on several benchmarks; and 
(iv) obtaining state-of-the-art performance on two Romanian ASR datasets.

\section{Lateral Inhibition}

\label{sec:li}


Inspired by the human brain's biological process of lateral inhibition, the neural lateral inhibition layer has been successfully applied in named entity recognition \cite{pais2022lateralinhib}. This process accounts for exciting neurons reducing the activity of neighboring neurons in the human brain \cite{Cohen2011}. Also, it provides an increased perception of the visual cortex under challenging scenarios, such as low-lighting conditions \cite{avramMP22}. Intuitively, we envisage that the new layer should be able to better focus on the actual voice data while possibly removing unwanted noise.

Following the original formulation~\cite{pais2022lateralinhib}, the lateral inhibition layer is described as follows:

\begin{equation}
\label{eq:matrices}
F(x)=x \cdot Diag(\Theta(x \cdot ZeroDiag(W) + b ))
\end{equation}
where $x$ is the input vector of the layer (i.e., the embedding representation produced by the RoWav2Vec2.0-VP-100k-LI model), $Diag(\cdot)$ denotes a diagonal matrix having the diagonal set to the vector presented as a parameter, $ZeroDiag(\cdot)$ generates a matrix with the zero value on the diagonal, $W$ is the weight matrix, $b$ corresponds to the bias values, and $\Theta(\cdot)$ is the Heaviside function (see Equation \ref{eq:heaviside}).

\begin{equation}
\label{eq:heaviside}
\Theta(x) = \left\{
\begin{matrix}
	1, x > 0  \\
	0, x \leq 0
\end{matrix}
\right.
\end{equation}

Following the analogy with the biological process, the Heaviside function determines which values can pass to the next layer. The decision is based on the adjacent values in the supplied embedding representation.
Equation \ref{eq:matrices} is used for the forward pass, with the Heaviside function included, thereby providing a strict pass or reject functionality for the input values. However, in the backward pass, the non-differentiable Heaviside function is replaced with the parameterized sigmoid function \cite{Wunderlich2021} (see Equation \ref{eq:sigmoid}, where $k$ is the scaling parameter). This technique, known as surrogate gradient learning \cite{8891809}, allows using a known derivative (see Equation \ref{eq:sigmoid_deriv}) in the backward pass.
\begin{equation}
\label{eq:sigmoid}
\sigma(x) = \frac{1}{1+e^{-kx}}
\end{equation}

\begin{equation}
\label{eq:sigmoid_deriv}
\sigma'(x) = k\sigma(x)\sigma(-x)
\end{equation}

\section{Experimental Settings}

\begin{table*}
    \centering

    \caption{Results of our models with or without lateral inhibition, which are trained on each dataset subset and evaluated on RSC, SSC, and RTASC, compared with other Romanian ASR models.}
    
        \begin{tabular}{l|c|cc|cc|cc}
            \toprule
             \multirow{ 2}{*}{\textbf{Model}} & \textbf{Train Size} &  \multicolumn{2}{c|}{\textbf{\underline{RSC}}} & \multicolumn{2}{c|}{\textbf{\underline{SSC}}} & \multicolumn{2}{c}{\textbf{\underline{RTASC}}} \\
              & \textbf{(\#hours)} & \textbf{WER(\%)} & \textbf{CER(\%)} & \textbf{WER(\%)} & \textbf{CER(\%)} & \textbf{WER(\%)} & \textbf{CER(\%)} \\ 
             \midrule
             DeepSpeech2 \cite{avram2020towards} & 230h & 5.34 & 1.59 & 25.55 & 9.617 & 37.21 & 17.66 \\
             TDNN-LSTM \cite{georgescu2019kaldi} & 235h & 4.55 & - & 19.25 & - & - & - \\ 
             CNN-TDNN \cite{georgescu2019kaldi} & 235h & 3.44 & - & \textbf{16.00} & - & - & - \\
             TDNN \cite{georgescu2019kaldi} & 235h & 3.32 & - & 16.85 & - & - & - \\
             TDNN-RNN \cite{georgescu2019kaldi} & 235h &  2.79 & - & 16.63 & - & - & - \\
             \midrule
             RoWav2Vec2.0-VP-100k-S & 10m & 44.78 & 9.13 & 68.40 & 22.55 & 80.73 & 33.68 \\
             RoWav2Vec2.0-VP-100k-LI-S & 10m & 35.00 & 7.05 & 58.05 & 18.72 & 76.33 & 31.47 \\
             \midrule
             
             RoWav2Vec2.0-VP-100k-M & 1h & 16.55 & 3.44 & 39.86 & 12.75 & 58.47 & 24.08 \\
             RoWav2Vec2.0-VP-100k-LI-M & 1h &  13.92 & 3.07 & 38.55 & 12.33 & 54.98 & 23.42 \\
             \midrule
             
             RoWav2Vec2.0-VP-100k-L & 10h & 4.80 & 1.62 & 28.23 & 11.55 & 44.52 & 23.04 \\
             RoWav2Vec2.0-VP-100k-LI-L & 10h & 3.95 & 1.18 & 24.73 & 9.35 & 37.12 & 16.50\\
             \midrule
             
            RoWav2Vec2.0-VP-100k-XL & 100h & 2.31 & 0.86 & 23.12 & 9.62 & 33.35 & 17.08 \\
            RoWav2Vec2.0-VP-100k-LI-XL & 100h & 1.80 & \textbf{0.70} & 19.21 & 7.96 & 29.69 & 13.95 \\
             \midrule
             
             RoWav2Vec2.0-VP-100k-XXL & 300h & 2.01 & 0.72 & 20.04 & 7.89 & 31.51 & 14.12 \\
             RoWav2Vec2.0-VP-100k-LI-XXL & 300h & \textbf{1.78} & 0.71 & 18.87 & \textbf{7.87} & \textbf{29.64} & \textbf{13.71} \\
             
             \bottomrule
        \end{tabular}
    \label{tab:wav2vec2_results}
\end{table*}

\label{sec:exp_setup}

\subsection{Dataset}

The fine-tuning of the RoWav2Vec2.0-VP-100k-LI model was done on a speech dataset whose composition contained ten Romanian corpora with transcribed audio files. The corpora contain recordings from several domains, including Wikipedia, News, Internet, and Legal. The resulting dataset has approximately 300 hours of transcribed speech from 222.7k utterances. It is composed of both reading and spontaneous speech, distributed in an imbalanced manner, with 229 hours of reading and 71 hours of spontaneous speech, respectively.

We further split our Romanian speech dataset into five subsets based on the total recording time by random sampling without replacement audio files until the desired size was reached: Small (S) - 10 minutes, Medium (M) - 1 hour, Large (L) - 10 hours, Extra Large (XL) - 100 hours, and Extra Extra Large (XXL) - the whole dataset. The split was necessary to evaluate the lateral inhibition performance in more extreme settings, i.e., with fewer labeled audio files.

\subsection{Fine-tuning}

We used the primary fine-tuning mechanism for the Wav2Vec 2.0 model as introduced in the original paper \cite{baevski2020wav2vec}. Therefore, using the raw audio input, we project the contextualized embeddings $c_i$ obtained by the model for each time step $i$ into a tensor $y_i$ whose dimensions match the number of letters in the Romanian alphabet, plus the space character and the blank token. We project the data using either the standard fully-connected layer or the lateral inhibition layer followed by a dense layer. Using the connectionist temporal classification algorithm \cite{graves2006connectionist}, we computed the loss between the predicted logits and target labels. We set $ k=10 $ for the lateral inhibition layer, which we believe is a good enough approximation of the surrogate gradient of the Heaviside function.
 
We employed the Adam method \cite{kingma2017adam} to optimize the loss with a learning rate set to $3e-5$ and a weight decay to $5e-3$. We fine-tuned each model on two NVIDIA 1080 TI GPUs. Due to GPU memory limitations, we set the batch size to 4 with a gradient accumulation of 8. 
In addition, we clipped the gradients from the back-propagation algorithm to 2 to improve training stability \cite{pascanu2013difficulty}.

\section{Results}

\label{sec:results}

\subsection{Romanian ASR}

We evaluate our models, namely RoWav2Vec2.0-VP-100k (i.e., without lateral inhibition) and RoWav2Vec2.0-VP-100k-LI (i.e., with lateral inhibition), on the test set of three corpora: Spontaneous Speech Corpus (SSC) \cite{georgescu2021improvements},  RSC, and RTASC. Compared with previous works on Romanian ASR, the results of the evaluation regarding WER and character error rate (CER) are listed in Table \ref{tab:wav2vec2_results}. In all our experiments, the decoding phase employs a 4-gram KenLM language model \cite{heafield2011kenlm} trained on the textual part of the corpus for contemporary Romanian language \cite{tufics2019little}.

Our model with lateral inhibition, trained on the full dataset (i.e., RoWav2Vec2.0-VP-100k-LI-XXL), obtains state-of-the-art performance on the RSC and RTASC corpora, achieving 1.78\% WER and 29.64\% WER, respectively\footnote{The high difference in WER between the two corpora comes from the type of utterances found in them: RSC contains common Romanian words and phonemes, while RTASC has more specific utterances from technology, with many words and phonemes borrowed from the English language.}. It improves the performance of the best Kaldi \cite{povey2011kaldi}-based ASR system, the Time Delay Neural Network - Recurrent Neural Network (TDNN-RNN) \cite{georgescu2019kaldi}, by 1.01\% WER on RSC and also the performance of the Romanian DeepSpeech2 model \cite{avram2020towards} on RTASC by 7.57\% WER.

However, our proposed models do not improve the performance on the SSC evaluation set, with our best variant (i.e., RoWav2Vec2.0-VP-100k-LI-XXL) falling behind 2.24\%  WER compared to the TDNN-RNN architecture. The main driver behind this difference is the need for more spontaneous speech data within our training corpus compared to the dataset used for training the state of the art. Specifically, the TDNN - Long Short-Term Memory (TDNN-LSTM), the Convolutional Neural Network - TDNN (CNN-TDNN), the TDNN, and the TDNN-RNN were all trained on a dataset with 235 hours of speech, namely 95 hours of read speech data from RSC and 140 hours of dedicated internal spontaneous speech data, similar to the one used in the SSC evaluation set. Meanwhile, we used only 71 hours of spontaneous speech data, approximately half the amount used to train the TDNN-based models. 

On the other hand, we increased the number of read speech data by decreasing the amount of spontaneous speech data within our training corpus. Hence, the performance of our best variant on the RSC evaluation set may have benefited from this fact. However, RoWav2Vec2.0-VP-100k-LI-XL still achieves almost state-of-the-art performance with 1.80\% WER on RSC, indicating that our methodology has not benefited too much from the increased amount of read speech data on this test set.

\begin{figure*}
    \centering
    \includegraphics[width=1\textwidth]{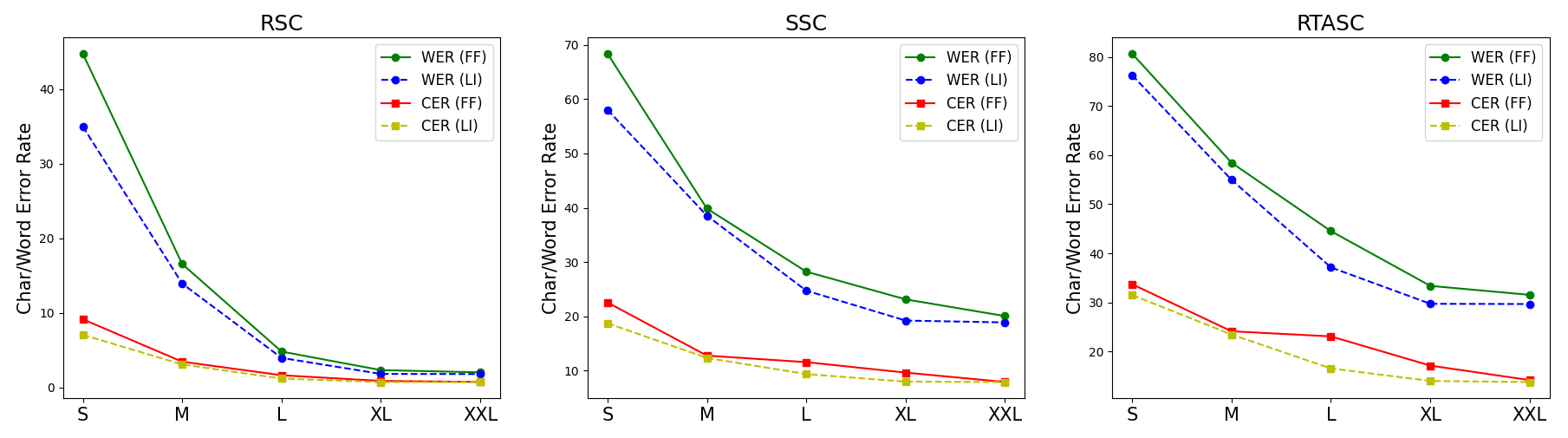}
    \caption{WER and CER comparison between fine-tuning the RoWav2Vec2.0-VP-100k model with a feed-forward (FF) layer or a lateral inhibition (LI) layer on each evaluation corpus subset.}
    \label{fig:comparison}
\end{figure*}

Apart from our best model, the rest of the variants performed reasonably well on each evaluation task, given the low amount of available training data. The RoWav2Vec2.0-VP-100k model obtained good results when fine-tuned on the L, XL, and XXL subsets, but the word error rate rapidly increased when the training dataset was switched to the more extreme cases (i.e., the M and S subsets). For instance, on the RSC dataset, the variants fine-tuned on the L, XL, and XXL subsets maintained a fairly good performance, achieving 4.80\%, 2.31\%, and 2.01\% WER, respectively (or 3.95\%, 1.80\%, and 1.78\% WER, respectively, with the lateral inhibition layer). However, the WER increased by more than three times on the RSC M subset and more than eight times on the RSC S subset, with our model obtaining 16.55\% and 44.78\% WER, respectively (or 13.92\% and 35.00\% WER with the lateral inhibition layer).

\subsection{Lateral Inhibition Layer Improvements}

We further analyze the improvements brought by the lateral inhibition in the RoWav2Vec2.0-VP-100k-LI models on all five evaluation subsets. An illustration of the difference in performance obtained by our model fine-tuned on all subsets is depicted in Figure \ref{fig:comparison}. We observe that the lateral inhibition layer decreases the error rates of the RoWav2Vec2.0-VP-100k-LI models in all our experiments. We also notice that, on average, the improvements become more significant for the smaller subsets. We believe this results from the increased regularization when the lateral inhibition layer is employed, mainly because it allows the model to focus better on the features of the actual human voice, thereby learning to distinguish the speech from the noise better even when the data is scarce.

\begin{figure}
    \centering
    \includegraphics[width=0.45\textwidth]{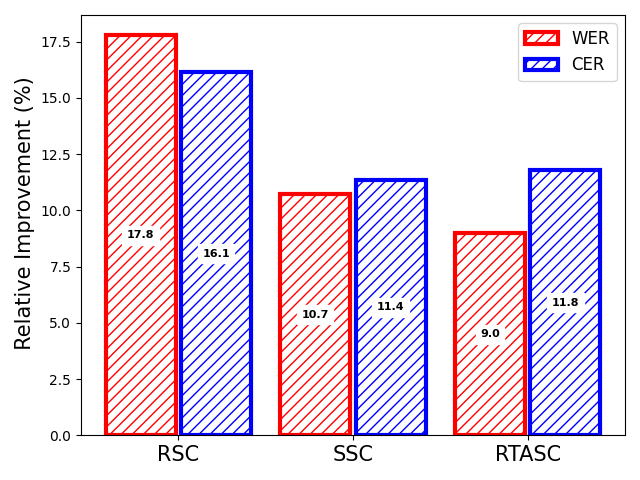}
    \caption{The average relative WER and CER improvement brought by the lateral inhibition layer.}
    \label{fig:rel_improv}
\end{figure}

We also compute the average relative improvement of the lateral inhibition mechanism to all the RoWav2Vec2.0-VP-100k-LI  variants on each evaluated corpus. We depict the results in Figure \ref{fig:rel_improv}. The greatest improvements are achieved on the RSC evaluation subsets, the lateral inhibition layer reducing the WER on average by 17.8\% and the CER by 16.1\%. The lowest average WER improvement (i.e., 9.0\%) is obtained on the RTASC evaluation subsets. Also, the lowest CER improvement (i.e., 11.4\%) is obtained on the SSC evaluation subsets. The average improvement over all evaluation subsets is 12.5\% for WER and 13.1\% for CER.


\section{Conclusions}
\label{sec:conclusion}

Automatic speech recognition for low-resource languages remains an important research direction. In this work, we applied the recently introduced mechanism, namely the lateral inhibition layer, which helps the speech recognition neural networks to better distinguish between the human voice and the surrounding noise. We performed experiments on the Romanian language using the RoWav2Vec2.0-VP-100k-LI models and a custom dataset composed of 300 hours of speech. The results showed that the lateral inhibition layer reduces, on average, the WER by 12.5\% over all the evaluated test sets. Furthermore, we achieved state-of-the-art performance on the RSC and RTASC datasets using this mechanism, obtaining 1.78\% WER and 29.64\% WER, respectively.

Future work considers experimenting with the lateral inhibition layer on languages other than Romanian and an evaluation of a speech dataset containing more than 300 hours. In addition, we intend to fine-tune other variants of the Wav2Vec 2.0 model, pre-trained on various datasets and with different methodologies, to validate that our results generalize beyond the pre-trained variant employed in this work.

\section*{Acknowledgements}
The research has been funded by the University Politehnica of Bucharest through the PubArt program.


\bibliographystyle{IEEEtran}
\bibliography{mybib}
%

\end{document}